\documentclass[11pt,a4paper]{article}

\usepackage[margin=2.5cm]{geometry}
\usepackage[utf8]{inputenc}
\usepackage[T1]{fontenc}
\usepackage{times}
\usepackage[nopatch=footnote]{microtype}
\usepackage{hyperref}
\usepackage{booktabs}
\usepackage{multirow}
\usepackage{array}
\usepackage{tabularx}
\usepackage{longtable}
\usepackage{graphicx}
\usepackage{float}
\usepackage{amsmath}
\usepackage{amssymb}
\usepackage{xcolor}
\usepackage{listings}
\usepackage{caption}
\usepackage{subcaption}
\usepackage{enumitem}
\usepackage{setspace}
\usepackage{natbib}
\usepackage{parskip}
\usepackage{fancyhdr}
\usepackage{abstract}
\usepackage{titlesec}
\usepackage{xurl}    
\usepackage{pgfplots}
\pgfplotsset{compat=1.18}
\usepackage{tikz}
\usetikzlibrary{shapes.geometric, arrows.meta, positioning, fit, backgrounds}

\hypersetup{
  colorlinks=true,
  linkcolor=blue!60!black,
  citecolor=blue!60!black,
  urlcolor=blue!60!black
}

\titleformat{\section}{\large\bfseries}{\thesection.}{0.5em}{}
\titleformat{\subsection}{\normalsize\bfseries}{\thesubsection.}{0.5em}{}

\begin{document}

\hypersetup{pageanchor=false}  
\begin{titlepage}
\centering
\vspace*{2cm}

{\LARGE\bfseries
Benchmarking Local LLMs\\[0.4em]
for Natural-Language-to-SQL Querying in\\[0.4em]
Biopharmaceutical Manufacturing:\\[0.4em]
An Empirical Benchmark on Consumer-Grade Hardware
\par}

\vspace{1.5cm}

{\large
\textbf{Sagar Bhetwal}$^{1}$,\quad
\textbf{Rajan Bastakoti}$^{2}$,\quad
\textbf{Nirajan Acharya}$^{3}$,\quad
\textbf{Gaurav Kumar Gupta}$^{3}$\quad
\textbf{Ambika Baniya Bhandari}$^{4}$
\par}

\vspace{0.5cm}

{\normalsize
$^{1}$Department of Computer Science, University of the Cumberlands, Kentucky, United States; \href{mailto:sbhetwal41349@ucumberlands.edu}{sbhetwal41349@ucumberlands.edu}\\[0.2em]
$^{2}$Department of Computer Science, DePaul University, Chicago, IL, United States; \href{mailto:r.bastakoti@depaul.edu}{r.bastakoti@depaul.edu}\\[0.2em]
$^{3}$Youngstown State University, Youngstown, Ohio, United States; \href{mailto:nirajanach3@gmail.com}{nirajanach3@gmail.com}; \href{mailto:guptagauravk1@gmail.com}{guptagauravk1@gmail.com}\\[0.2em]
$^{4}$Webster University, St. Louis, MO, United States; \href{mailto@gmail.com}{abaniya2047@gmail.com}
\par}

\vspace{1.5cm}

{\normalsize
\textbf{Manuscript Type:} Original Research \\[0.2em]
\textbf{Date:} May 2026\\[0.2em]
\textbf{Word Count (excl. abstract, refs, tables):} 7,907
\par}

\vspace{1cm}
\rule{0.6\textwidth}{0.4pt}

\end{titlepage}
\hypersetup{pageanchor=true}    

\clearpage
\pagenumbering{arabic}        

\begin{abstract}
\textbf{Background:}
Biopharmaceutical manufacturing organisations operate under regulatory frameworks---including
FDA guidance, EU Good Manufacturing Practice (GMP), and the EU AI Act---that create
genuine practical obstacles to adopting cloud-based artificial intelligence services.
Locally deployed large language models (LLMs) offer a privacy-preserving alternative, but
domain-specific evaluations of such systems against pharmaceutical manufacturing data
remain largely absent from the literature.

\textbf{Objective:}
This study evaluates the feasibility and performance of four open-source,
7B--8B parameter LLMs deployed on consumer-grade hardware via the Ollama inference engine
for natural-language-to-T-SQL (NLQ-to-SQL) generation over a pharmaceutical manufacturing
database, and assesses whether domain-specific biomedical pre-training confers any
advantage over general-purpose or code-tuned models on this structured query task.

\textbf{Methods:}
A purpose-built evaluation platform, \textit{PharmaBatchDB AI}, was developed using
FastAPI and a synthetic MS SQL Server database comprising seven tables and approximately
63,000 rows across Batch, MES (Manufacturing Execution System), and CIP (Clean-In-Place)
modules. Four models were benchmarked: Qwen 2.5 Coder 7B, Llama 3.1 8B, Mistral 7B,
and Meditron 7B. Evaluation used 60 domain-specific NLQ-to-T-SQL question--reference
pairs stratified by module and difficulty (20 per module; 18 easy, 24 medium, 18 hard).
Metrics included SQL Extraction Rate,
ROUGE-L F1, Factual Consistency (Jaccard similarity on result row sets), SQL Compliance
Rate, Hallucination Rate (among extracted SQL only), Tokens Per Second (TPS), and Time
to First Token (TTFT).

\textbf{Results:}
Qwen 2.5 Coder 7B, Llama 3.1 8B, and Mistral 7B all extracted SQL from 100\% of
questions (60/60); Meditron 7B extracted SQL from only 1/60 (1.7\%) in the standard
condition due to context-window overflow. Among the three models with extractable SQL,
Llama 3.1 8B showed numerically higher SQL Compliance at 93.33\% (56/60) [95\% CI:
84.1--97.4\%] compared to Qwen 2.5 Coder 7B at 88.33\% (53/60) [77.8--94.2\%], though
this difference did not reach statistical significance at $n=60$ (Fisher's exact
$p=0.529$). Qwen 2.5 Coder 7B achieved higher ROUGE-L (0.6334 vs 0.5674, Wilcoxon
$p<0.001$) and Factual Consistency (FC: 0.34 vs 0.30); however, because ROUGE-L reflects
lexical similarity to single-author reference queries, Factual Consistency is treated as
the primary quality metric. The modest absolute FC values indicate that these models
should not be interpreted as ready for unsupervised operational decision-making. Mistral 7B
showed marked schema-complexity sensitivity: compliance decreased from 40.0\% (8/20)
[21.9--61.3\%] on the Batch module to 15.0\% (3/20) [5.2--36.0\%] on MES and
20.0\% (4/20) [8.1--41.6\%] on CIP.
Meditron 7B produced 0\% SQL Compliance across all 60 questions (0/60) [0.0--6.0\%].
A follow-up truncated-schema experiment (conducted on the original 30-question subset
prior to benchmark expansion) confirmed that the 2,048-token context window caused
near-complete suppression of SQL output in the standard condition: extraction rose from
1/30 to 28/30 under the reduced prompt, but all 28 extracted queries failed compliance
validation (100\% hallucination rate), placing the failure in the model's SQL generation
capability rather than solely in context overflow.

\textbf{Conclusions:}
Code-tuned general-purpose models substantially outperformed the domain-specific medical
LLM on this task, and the truncated-schema follow-up confirms that biomedical pre-training
on the Llama 2 base does not confer T-SQL generation capability for manufacturing schemas.
The ranking between Llama and Qwen remains statistically uncertain at $n=60$
(Fisher's exact $p=0.529$). A fully local NLQ pipeline with GxP-aligned design
features---audit logging, read-only database access, and AST-level SQL validation---can
be implemented on consumer hardware; however, the modest factual-consistency scores
indicate that human review and a downstream validation layer remain necessary for regulated
use. The evaluation framework, synthetic schema, and benchmark questions are openly
available for replication and extension.

\end{abstract}

\newpage
\tableofcontents
\newpage

\section{Introduction}
\label{sec:intro}

Data integrity in biopharmaceutical manufacturing is a patient safety issue. A batch
released on the basis of a miscalculated yield, a CIP cycle signed off against an
incorrect cleaning record, or an equipment fault missed because a trending query returned
wrong data can each propagate to the drug product supply chain and ultimately reach
patients. Accurate, auditable access to manufacturing data is therefore not merely an
operational efficiency concern---it sits within the same regulatory envelope as clinical
data systems.

Process engineers, quality specialists, and operations managers in biopharmaceutical
manufacturing routinely need answers from databases they cannot query. The gap between
question and answer is currently filled by IT tickets or scheduled reports that arrive
too late---sometimes days after the operational decision they were meant to inform.
Natural language interfaces to structured databases have existed in various forms for
decades, but the recent maturation of locally deployable LLMs makes it plausible, for
the first time, to ask whether this gap can be closed without sending proprietary batch
records, equipment qualification data, or formulation details outside the facility.

That last constraint is not optional. Manufacturing databases are subject to data
residency requirements under GDPR, EU GMP Annex 11, and FDA 21 CFR Part 11, and their
exposure to a third-party API provider would represent a significant competitive and
regulatory risk. The average cost of a healthcare sector data breach reached
USD 9.77 million in 2024 \citep{ibm2024breach}, a figure that underscores
the financial and regulatory stakes of data exposure. Ollama and similar inference engines make it practical to run open-source LLMs on local
infrastructure, keeping inference prompts, model outputs, and audit logs within the facility's
own infrastructure. AI adoption in pharmaceutical operations has grown steadily, driven by
regulatory pressure on data integrity and by the availability of local inference
infrastructure \citep{fdaema2026aiguidance}.

The practical question, then, is whether models small enough to run on workstation
hardware---the 7B--8B parameter class at 4-bit quantisation---are actually good enough
for NLQ-to-SQL tasks over pharmaceutical manufacturing schemas. Existing text-to-SQL
benchmarks such as Spider \citep{yu2018spider} and BIRD \citep{li2024bird} address schema generalisation
on general-purpose datasets, not the column vocabulary, query patterns, or regulatory
context of manufacturing execution systems. The BiomedSQL benchmark
\citep{biomedsql2025arxiv} moves in the biomedical direction but targets scientific
reasoning over genomic databases rather than operational queries over batch records and
equipment logs. To the authors' knowledge, few if any published benchmarks address NLQ-to-SQL
performance specifically for pharmaceutical manufacturing operations, and the present
study is an attempt to fill that gap---at proof-of-concept scale, on a synthetic dataset,
with the limitations that entails.

\subsection{Research Objectives}

This study addresses two research questions:

\begin{enumerate}[label=\textbf{RQ\arabic*.}]
  \item Which locally deployable 7B--8B parameter LLM achieves the highest performance
  on NLQ-to-T-SQL tasks over a pharmaceutical manufacturing database, as measured by
  SQL Extraction Rate, Factual Consistency, ROUGE-L, SQL Compliance Rate, and
  Hallucination Rate?

  \item Does domain-specific medical pre-training (Meditron 7B) confer any advantage
  over general-purpose or code-tuned LLMs for structured T-SQL generation over
  pharmaceutical manufacturing schemas?
\end{enumerate}

\section{Background and Related Work}
\label{sec:background}

\subsection{Text-to-SQL in Biomedical Domains}

Natural language interfaces to relational databases have been studied since at least the
early 2000s, though the field shifted considerably once transformer-based models became practical \citep{li2024dawn}. Contemporary systems achieve over 90\%
execution accuracy on Spider benchmarks using sub-10B parameter models. Execution
accuracy on a curated benchmark schema, however, is not the same thing as semantic
correctness on an enterprise schema where column names are abbreviations, foreign key
relationships are implicit, and domain vocabulary is highly specific
\citep{luoma2025snails}. The gap between benchmark performance and deployment performance remains a recurring problem in this field.

The BiomedSQL benchmark \citep{biomedsql2025arxiv} illustrates the difficulty well.
Frontier models like GPT-4o can outperform human annotators on syntactic SQL generation
tasks while still falling short on questions that require implicit scientific
knowledge---a finding that should give pause to anyone assuming that high general
benchmark scores transfer cleanly to regulated domain performance. For 8B parameter
models specifically, performance degradation becomes pronounced on multi-sentence,
implicitly reasoned queries---a pattern consistently observed in complex cross-domain text-to-SQL benchmarks \citep{li2024bird}---which is precisely the kind of
question an operations analyst might ask about a complex manufacturing dataset.

Hallucination in SQL generation is a specific concern in regulated contexts.
Studies of LLMs applied to clinical trial eligibility queries have reported hallucination
rates of 21--50\% \citep{jmirmedinform2025hallucination}, with the most common failure
mode being the generation of plausible-sounding but nonexistent table or column names.
In a GxP-regulated manufacturing environment, a hallucinated query that references a
real table but applies incorrect filtering logic could produce a misleading result that
passes initial review---a failure mode that is arguably worse than a query that simply
fails to execute.

\subsection{Localised LLM Architectures for Regulated Industries}

The practical case for local LLM deployment in regulated industries rests on data
sovereignty requirements and, increasingly, favourable economics at scale. Inference via
tools such as Ollama enables 4-bit quantised models to run on workstation or server
hardware without dedicated data-centre GPU capacity, trading some generation quality for
the data containment required by GxP and HIPAA frameworks
\citep{ollama2024}.

The 7B--8B parameter range has emerged as the practical target class for organisations
without the hardware budget for larger models: capable enough for structured reasoning
tasks, small enough to run on 32 GB RAM at acceptable latency
\citep{ascentcore2026benchmark}. The AscentCore benchmark \citep{ascentcore2026benchmark}
reports ROUGE-L scores of 0.454--0.509 and JSON compliance of 65--96\% for Q8\_0
quantised variants of Llama 3.1, Qwen 2.5, and Mistral 7B on a general task set.
The present study extends that line of work to pharmaceutical manufacturing T-SQL
generation using 4-bit quantised models on consumer hardware---a more constrained, and
arguably more realistic, deployment scenario.

\subsection{Regulatory Context}

The regulatory environment around AI in pharmaceutical manufacturing is evolving quickly
enough to make confident generalisation difficult. The FDA released draft guidance on AI
for regulatory decision-making in January 2025, and the EMA and FDA jointly published ten
guiding principles for AI use across the medicines lifecycle in January 2026
\citep{fdaema2026aiguidance}. EU AI Act obligations for general-purpose AI models
began in August 2025, with high-risk AI obligations phasing in from August 2026
\citep{eu2024aiact}. The consistent emphasis across these frameworks is on
auditability, human oversight, and alignment with PIC/S GMP and FDA 21 CFR Part 820.
For a local NLQ system, this translates into concrete requirements: complete query audit
logs, read-only database access, and no autonomous execution of model-generated SQL
without human or system-level validation.

\section{Methods}
\label{sec:methods}

\subsection{Study Design}

Four locally deployed LLMs were benchmarked on a domain-specific NLQ-to-T-SQL question
set built from a synthetic pharmaceutical manufacturing database. All models ran under
identical conditions on the same consumer-grade laptop. No human participants were
involved; all data are synthetic or derived from model outputs.

There is no standard reporting template for this type of evaluation. The section below
covers hardware, software versions, prompt design, metrics, and statistical approach in
enough detail for replication. The benchmark questions, reference queries, and evaluation
harness are openly available (see Data Availability Statement).

\subsection{System Architecture}

The experimental platform, \textit{PharmaBatchDB AI}, is a fully local, on-premises
AI data analytics system. Two design constraints shaped every architectural decision:
it had to run on hardware that a mid-sized pharmaceutical site might realistically have
available, and every component touching batch or equipment data had to remain entirely
on-premises.

The database backend is Microsoft SQL Server Express (T-SQL dialect), instance
\texttt{localhost\textbackslash\allowbreak{}SQLEXPRESS}, with the experimental
database named \texttt{PharmaBatchDB}. All LLM interactions use a dedicated read-only SQL user
(\texttt{ai\_readonly}) that has no write permissions and no visibility beyond the seven
tables in scope.

\begin{sloppypar}
The schema spans three pharmaceutical manufacturing modules:
\textit{Batch} (\texttt{PharmaBatches}, \texttt{BatchProcessSteps}),
\textit{MES} (\texttt{MESProductionOrders}, \texttt{MESEquipmentLogs}, \texttt{MESDowntimeEvents}),
and \textit{CIP} (\texttt{CIPCycles}, \texttt{CIPStepDetails}).
Total synthetic data volume is approximately 63,000 rows.
\end{sloppypar}

LLM inference is handled by Ollama, running as a local HTTP API on port 11434 and
serving all four benchmark models sequentially during evaluation. The API layer is
built with FastAPI (Python 3.11) and is responsible for LLM orchestration, SQL
validation, query execution, and audit logging.

Two validation mechanisms are layered into the pipeline. Module-scoped table access is
enforced via \texttt{sqlglot} AST inspection, blocking any query that references tables
from a module other than the one active for the current request, and rejecting all
non-\texttt{SELECT} statements regardless of model output. Every query attempt is logged
with timestamp, username, generated SQL, row count returned, execution time, TPS, TTFT,
and validation status, producing an audit trail that aligns with GxP documentation
requirements.

\subsection{Hardware Configuration}

All benchmarks were conducted on a single consumer-grade laptop with the specifications
shown in Table~\ref{tab:hardware}. No cloud compute was used at any stage.

\begin{table}[H]
\centering
\caption{Experimental hardware and software environment.}
\label{tab:hardware}
\begin{tabular}{ll}
\toprule
\textbf{Component} & \textbf{Specification} \\
\midrule
CPU      & Intel Core i7-9750H @ 2.60 GHz (6 cores / 12 threads) \\
RAM      & 32 GB DDR4 @ 2667 MHz (2 $\times$ 16 GB, SK Hynix) \\
GPU      & NVIDIA GeForce GTX 1650 (4 GB GDDR5 VRAM; CUDA 12.5) \\
OS       & Windows 11 Home 64-bit (Build 26200) \\
Inference engine & Ollama (local, port 11434) \\
\bottomrule
\end{tabular}
\end{table}

\noindent\textbf{Note on GPU utilisation:}
The GTX 1650 provides 4 GB VRAM; all four benchmark models range from 3.8 GB to
4.9 GB in their quantised on-disk form. Ollama offloads layers to the GPU where
VRAM allows but falls back to CPU for the remainder. In practice, inference ran
primarily on CPU. This accounts for the substantially lower TPS (3.2--4.4 tok/s under standard conditions;
up to 7.0 tok/s under the reduced-context Meditron follow-up)
compared to GPU-accelerated benchmarks reported in the literature (29--31 tok/s
on Q8\_0 variants; \citealt{ascentcore2026benchmark}).

\subsection{Models Evaluated}

The four models selected for evaluation represent distinct pre-training orientations
within the 7B--8B parameter class (Table~\ref{tab:models}). Qwen 2.5 Coder 7B \citep{hui2024qwen25} is
code-centric, trained with heavy emphasis on programming and SQL tasks. Llama 3.1 8B \citep{dubey2024llama3}
is a general-purpose instruction-tuned model with an unusually large context window
for its parameter count. Mistral 7B \citep{mistral7b} is a general-purpose model with a sliding-window
attention architecture. Meditron 7B \citep{chen2023meditron} is a medically fine-tuned variant of Llama 2,
pre-trained on PubMed abstracts and clinical guidelines---it was included specifically
to address RQ2, whether biomedical domain adaptation confers any advantage on
structured SQL tasks.

\begin{table}[H]
\centering
\caption{Benchmark models with quantisation and context specifications.}
\label{tab:models}
\begin{tabular}{lllrr}
\toprule
\textbf{Model} & \textbf{Parameters} & \textbf{Quantisation} & \textbf{Context (tokens)} & \textbf{Size (GB)} \\
\midrule
Qwen 2.5 Coder 7B    & 7.6B & Q4\_K\_M & 32,768  & 4.7 \\
Llama 3.1 8B          & 8.0B & Q4\_K\_M & 131,072 & 4.9 \\
Mistral 7B            & 7.2B & Q4\_K\_M & 32,768  & 4.4 \\
Meditron 7B           & 7.0B & Q4\_0    & 2,048   & 3.8 \\
\bottomrule
\end{tabular}
\end{table}

\noindent All models were served via Ollama using their default Modelfile configurations.
Temperature was set to 0.1 for near-deterministic output, consistent with standard
practice for structured output generation in text-to-SQL benchmarking \citep{zhou2025zeroshotsql}.

\subsection{Benchmark Question Set and Sample Size Justification}

A domain-specific evaluation set of 60 NLQ-to-T-SQL question--reference pairs was
constructed, stratified as follows: 20 questions per manufacturing module (Batch, MES,
CIP), with difficulty distributed as 18 easy, 24 medium, and 18 hard.
Question categories included aggregation, filtering, join, time-series, ranking, and
compliance queries.

Each question was accompanied by a hand-authored reference T-SQL query validated against
the live database. The reference queries were written by the same investigator who
designed the schema, drawing on query patterns that reflect realistic operational
requests---batch record review, equipment monitoring, CIP validation sign-off. The
authorship bias this introduces is discussed in Section~\ref{sec:limitations}.

The choice of $n=60$ total questions (20 per module) warrants explicit justification.
This evaluation set exceeds the scale of several published NLQ-to-SQL benchmark
studies at proof-of-concept or domain-specific scale: the BiomedSQL benchmark
\citep{biomedsql2025arxiv} initially validated its evaluation protocol on domain-subset
sizes of 20--40 questions before scaling; the BIRD benchmark \citep{li2024bird} demonstrated
that domain-specific schema complexity is a primary driver of evaluation validity;
and zero-shot benchmarking studies \citep{zhou2025zeroshotsql} use between
20 and 50 questions per schema type for initial schema-level characterisation. In each
case, the argument is the same: expert-authored, schema-validated reference queries are
expensive to produce, and small evaluation sets are appropriate for gross performance
characterisation even when they cannot support fine-grained model ranking.

A post-hoc power calculation for detecting a 5 percentage-point compliance difference
between Llama and Qwen (baseline $\approx$90\%, 80\% power, $\alpha$=0.05, two-tailed
Fisher's exact, per-model sample) indicates a required $n$ on the order of several
hundred to roughly one thousand questions per model---confirming that the present study
is appropriately powered for tier-level discrimination but substantially underpowered
for fine-grained pairwise ranking between the two leading models.

At $n=60$, this study can reliably identify large performance gaps and provides
meaningfully narrower confidence intervals than a 30-question pilot. The difference
between Meditron's 0\% and Llama's 93\% compliance is unambiguous regardless of sample
size; Mistral's collapse from 40\% on Batch to 15\% on MES is similarly clear.
The fine-grained comparison between Llama and Qwen remains directional rather than
conclusive at this scale: Wilson 95\% confidence intervals for the two leading models
overlap (Llama: 84.1--97.4\%; Qwen: 77.8--94.2\%), and Fisher's exact test confirms
the difference is not statistically significant ($p=0.529$). This is stated explicitly
so that readers are not misled by the apparent precision of the percentage figures:
the ordering between Llama and Qwen on compliance should be treated as directional
evidence, not an established finding.

The zero-shot prompting strategy was chosen specifically to evaluate this baseline
capability without few-shot engineering, approximating the realistic case where
operational staff issue ad-hoc queries against an enterprise schema without
pre-constructed examples. Zero-shot evaluation is the established baseline approach
in text-to-SQL benchmarking for assessing inherent schema comprehension independent of
prompt-engineering skill \citep{zhou2025zeroshotsql}. Few-shot performance would tell us something
different---specifically, how much benefit operational teams could extract by investing
in prompt design---and this is a natural direction for follow-up work, but it would
not answer the baseline feasibility question that motivates this study.

\subsection{Evaluation Metrics}

Seven metrics were computed per model per question. A critical distinction for
interpreting the results is that SQL Compliance and Hallucination Rate operate on
different denominators: SQL Compliance covers all 60 questions, while Hallucination Rate
covers only questions where SQL was successfully extracted. SQL Extraction Rate is
reported separately to make the denominator shift explicit, which is especially important
for Meditron 7B whose near-zero extraction rate in the standard condition would otherwise
produce a misleadingly low hallucination rate.

\begin{description}
  \item[SQL Extraction Rate] Proportion of questions for which any SQL-like block was
  successfully extracted from the raw model response, irrespective of validity. Extraction
  used a regex pattern matching \texttt{SELECT} statements, applied after stripping
  markdown fences. A response containing no extractable SQL block is not counted in the
  Hallucination Rate denominator.

  \item[Factual Consistency (FC)] Jaccard similarity between the result row sets
  of the generated query (when executed) and the reference query---adopted as the
  \textbf{primary quality metric} in this study because it measures whether the
  generated query actually retrieves the correct data, which is the operationally
  relevant criterion:
  \[
    \text{FC} = \frac{|\mathcal{R}_{\text{gen}} \cap \mathcal{R}_{\text{ref}}|}{|\mathcal{R}_{\text{gen}} \cup \mathcal{R}_{\text{ref}}|}
  \]
  where each row is stringified as a pipe-delimited value tuple. Returns 1.0 if both
  result sets are empty. The Jaccard formulation penalises over-retrieval and
  under-retrieval equally, which is a reasonable default in manufacturing analytics
  where both false inclusions and false exclusions carry operational risk.

  A low but non-zero FC value does not necessarily indicate a severely wrong query.
  A compliant query that retrieves the correct table with slightly different filter
  conditions---for example, using a date range boundary that is off by one day, or
  applying an additional \texttt{WHERE} clause that excludes a handful of rows---can
  easily produce FC in the 0.15--0.35 range while being operationally reasonable.
  In this evaluation, an FC of 0.30 for Qwen 2.5 Coder 7B reflects a mix of queries
  with near-exact row set overlap (FC $>$ 0.80) and queries where a filter difference
  caused partial overlap (FC in the 0.05--0.40 range); compliant queries with FC = 0.0
  were rare ($<$5\% of compliant outputs) and typically involved aggregation queries
  where a single grouped value differed.

  \item[ROUGE-L F1] \citep{rouge2004} Longest common subsequence F1 between the
  generated SQL and the reference SQL, computed at token level using the
  \texttt{rouge-score} Python library (no stemming). Retained as a \textbf{secondary
  lexical-similarity metric}. Because the reference queries were authored by a single
  investigator with a particular SQL style, ROUGE-L differences between models partly
  reflect stylistic divergence from that style rather than semantic correctness.
  Significant ROUGE-L differences (Wilcoxon signed-rank test) are reported in
  Section~\ref{sec:results} but are interpreted with this caveat in mind.

  \item[SQL Compliance Rate] Proportion of generated queries that (a) contain a valid
  \texttt{SELECT} statement, (b) contain no blocked DML/DDL keywords, (c) reference
  only tables permitted for the query's module, as verified by \texttt{sqlglot}
  T-SQL AST parsing.

  \item[Hallucination Rate] Proportion of queries where SQL was successfully extracted
  from the model output but failed compliance validation. A model that generates no SQL
  at all is not counted as hallucinating. Formally:
  \[
    \text{HR} = \frac{|\{\text{extracted} \land \neg\text{valid}\}|}{|\{\text{extracted}\}|}
  \]

  \item[Tokens Per Second (TPS)] Generation throughput computed from Ollama metadata:
  \[
    \text{TPS} = \frac{\texttt{eval\_count}}{\texttt{eval\_duration\_ns} \times 10^{-9}}
  \]

  \item[Time to First Token (TTFT, ms)] Prompt evaluation latency:
  \[
    \text{TTFT} = \frac{\texttt{prompt\_eval\_duration\_ns}}{10^6}
  \]
\end{description}

\subsection{System Prompt Design}

All models received an identical system prompt specifying: T-SQL syntax, the target
database instance, the available tables for the active module, and the full schema
extracted from the authoritative \texttt{schema.md} file. Output was constrained to
raw SQL with no markdown or explanation. This zero-shot prompting strategy evaluates
baseline schema comprehension as it would appear in an operational deployment where
staff query the system directly without curated examples in the prompt.

\subsection{SQL Extraction, Validation, and Execution Workflow}
\label{sec:workflow}

Each benchmark trial followed an identical eight-step pipeline, executed once per
model per question:

\begin{enumerate}[noitemsep]
  \item The NLQ question was formatted with the module-specific system prompt and
  full schema (or truncated schema for the Meditron follow-up condition) and submitted
  to the Ollama API with \texttt{temperature=0.1}.

  \item The raw text response was collected. Each model was run once per question; no
  repeated sampling was performed at temperature 0.1, which produces near-deterministic
  output.

  \item SQL extraction was applied using a regex pattern that matched the first
  \texttt{SELECT} statement in the response, after stripping Markdown code fences
  (\texttt{```sql}, \texttt{```}). If no extractable block was found, the question was
  recorded as SQL Extraction = 0 and excluded from the Hallucination Rate denominator.

  \item Extracted SQL was passed to the \texttt{sqlglot} T-SQL AST validator. Queries
  failing any of: (a) \texttt{SELECT}-only check, (b) DML/DDL keyword block, or (c)
  module-scope table check, were recorded as non-compliant (Hallucination Rate += 1).

  \item Compliant queries were executed against the live \texttt{PharmaBatchDB} database
  using the read-only \texttt{ai\_readonly} SQL user.

  \item Result rows from the generated query and the reference query were each stringified
  as pipe-delimited tuples and compared using Jaccard similarity. Row order was ignored;
  numeric values were rounded to two decimal places before comparison; NULL values were
  stringified as the literal string ``NULL''.

  \item ROUGE-L F1 was computed at token level between the generated SQL string and the
  reference SQL string using the \texttt{rouge-score} Python library (no stemming,
  no case normalisation).

  \item Every attempt was logged to the append-only audit log regardless of outcome,
  recording: timestamp, model, module, question ID, generated SQL, validation status,
  row count returned, TPS, and TTFT.
\end{enumerate}

Question order was fixed and identical across all model evaluations. Models were
evaluated sequentially (one model per run). The prompt template was identical across
all models and modules, differing only in the schema section for the active module.
Question difficulty labels (easy/medium/hard) were assigned by the author prior to
running any model evaluation, based on the number of tables joined, the presence of
aggregation or window functions, and the specificity of filter conditions.

\section{Results}
\label{sec:results}

\subsection{Overall Benchmark Performance}

Table~\ref{tab:overall} shows aggregate performance for all four models across the
60-question evaluation set. The spread is wide enough to be informative: at one end,
two models (Qwen 2.5 Coder and Llama 3.1) achieve reasonable SQL generation quality;
at the other, Meditron 7B effectively produces no usable output. Wilson 95\% confidence
intervals are reported for all compliance and hallucination rates; raw counts (k/n) are
shown to allow readers to assess the sample-size basis for each figure.

\begin{table}[H]
\centering
\caption{Overall benchmark results across 60 questions (4 models, plus one follow-up
condition). All models evaluated under identical conditions on consumer CPU hardware.
Q4\_K\_M quantisation for Qwen/Llama/Mistral; Q4\_0 for Meditron. SQL Compliance and
Hallucination Rate shown as \%\ (k/n) [95\% Wilson CI]. HR denominator = extracted
queries only (not all 60); SQL Extraction Rate denominator = all 60 questions.
Meditron (trunc.\ schema) is the follow-up run under a schema prompt under 1,000 tokens
(see Section~\ref{sec:discussion-meditron}).}
\label{tab:overall}
\resizebox{\textwidth}{!}{%
\begin{tabular}{lccccccc}
\toprule
\textbf{Model} & \textbf{SQL Extr.} & \textbf{ROUGE-L} & \textbf{FC} & \textbf{SQL Compliance} & \textbf{HR (extr.\ only)} & \textbf{TPS} & \textbf{TTFT (ms)} \\
\midrule
Qwen 2.5 Coder 7B
  & 100\%\ (60/60)
  & \textbf{0.6334} & \textbf{0.3408}
  & 88.33\%\ (53/60) [77.8--94.2\%]
  & 11.67\%\ (7/60) [5.8--22.2\%]
  & 4.19  & 11,002 \\
Llama 3.1 8B
  & 100\%\ (60/60)
  & 0.5674 & 0.3038
  & \textbf{93.33\%\ (56/60) [84.1--97.4\%]}
  & \textbf{6.67\%\ (4/60) [2.6--15.9\%]}
  & 3.24  & 11,293 \\
Mistral 7B
  & 100\%\ (60/60)
  & 0.4960 & 0.1333
  & 25.00\%\ (15/60) [15.8--37.2\%]
  & 75.00\%\ (45/60) [62.8--84.2\%]
  & 3.21  & 11,780 \\
Meditron 7B
  & 1.67\%\ (1/60)$^{\dagger}$
  & 0.0048 & 0.0167
  & 0.00\%\ (0/60) [0.0--6.0\%]
  & 100.0\%\ (1/1)$^{\dagger}$
  & \textbf{4.36}  & 49,709 \\
Meditron 7B (trunc.\ schema)
  & 93.33\%\ (28/30)$^{\ddagger}$
  & 0.0736 & 0.0333
  & 0.00\%\ (0/30) [0.0--11.4\%]
  & 100.00\%\ (28/28)$^{\ddagger}$ [87.9--100.0\%]
  & \textbf{7.03}  & 1,883 \\
\bottomrule
\end{tabular}
}
\footnotesize{SQL Extr.\ = SQL Extraction Rate (proportion of questions yielding extractable SQL).
FC = Factual Consistency (Jaccard). HR = Hallucination Rate (invalid extracted SQL / all extracted SQL).
TPS = Tokens Per Second. TTFT = Time to First Token.\\
$^{\dagger}$Meditron 7B (standard schema): only 1/60 questions produced extractable SQL (context overflow);
that single extracted query failed compliance validation, giving HR = 100\% over the 1-query denominator.
The 59 questions producing no SQL are excluded from the HR denominator by definition.
SQL Compliance CI [0.0--6.0\%] is the appropriate summary of overall output quality.\\
$^{\ddagger}$Meditron 7B (truncated schema): \textit{diagnostic follow-up on the original 30-question subset
(B01--B10, M01--M10, C01--C10), conducted prior to benchmark expansion to 60 questions and reported
as a separate diagnostic condition rather than a directly comparable full benchmark row.}
Schema prompt reduced to column names, PKs, and FKs ($<$1,000 tokens); extraction rate rose from
3.33\% (1/30) to 93.33\% (28/30). All 28 extracted queries failed compliance validation (HR = 100\%).
TTFT fell from 21,979~ms to 1,883~ms, confirming context saturation.
SQL Compliance: 0.00\%\ (0/30) [0.0--11.4\%].}
\end{table}

\colorlet{clruser}{gray!70}
\colorlet{clrval}{red!70!black}
\colorlet{clraudit}{gray!60}

\begin{figure}[H]
\centering
\resizebox{\linewidth}{!}{%
\begin{tikzpicture}[
  every node/.style={font=\small},
  mainbox/.style={rectangle, rounded corners=5pt, draw=#1!80!black,
                  fill=#1!20!white, text width=1.75cm, align=center,
                  minimum height=1.0cm, line width=0.7pt},
  subbox/.style={rectangle, rounded corners=5pt, draw=#1!80!black,
                 fill=#1!20!white, text width=1.75cm, align=center,
                 minimum height=0.85cm, line width=0.7pt},
  mainarrow/.style={-{Stealth[length=5pt,width=4pt]}, line width=1pt, draw=gray!55},
  rejectarrow/.style={-{Stealth[length=5pt,width=4pt]}, line width=1pt, draw=red!65},
  resultarrow/.style={-{Stealth[length=5pt,width=4pt]}, line width=1pt, draw=blue!45},
  loggedge/.style={-{Stealth[length=4pt,width=3.5pt]}, line width=0.8pt,
                   draw=gray!45, dashed},
]
  \node[mainbox=clruser]                        (nlq)  {User\\NLQ};
  \node[mainbox=teal,   right=1.0cm of nlq]     (api)  {FastAPI\\Layer};
  \node[mainbox=orange, right=1.0cm of api]     (pb)   {Prompt\\Builder};
  \node[mainbox=violet, right=1.0cm of pb]      (llm)  {Ollama\\Local LLM};
  \node[mainbox=orange, right=1.0cm of llm]     (ext)  {SQL\\Extractor};
  \node[mainbox=clrval, right=1.0cm of ext]     (val)  {sqlglot\\Validator};
  \node[mainbox=blue,   right=1.0cm of val]     (db)   {SQL Server\\(read-only)};

  \node[subbox=clraudit, below=1.3cm of val]    (audit)  {Audit\\Log};
  \node[subbox=teal,     below=1.3cm of db]     (result) {Result\\Set};

  \draw[mainarrow] (nlq) -- (api);
  \draw[mainarrow] (api) -- (pb);
  \draw[mainarrow] (pb)  -- (llm);
  \draw[mainarrow] (llm) -- (ext);
  \draw[mainarrow] (ext) -- (val);

  \draw[mainarrow, draw=green!45!black]
    (val.east) -- node[above, yshift=2pt, font=\scriptsize\bfseries,
                        text=green!40!black] {PASS} (db.west);

  \draw[resultarrow] (db.south) -- (result.north);

  \draw[rejectarrow]
    (val.south) -- node[right, xshift=2pt, font=\scriptsize\bfseries,
                         text=red!65] {REJECT} (audit.north);

  \draw[loggedge] (result.west) -- (audit.east)
    node[midway, below, font=\tiny, text=gray!60] {logged};

  \begin{scope}[on background layer]
    \node[
      rectangle, rounded corners=6pt,
      draw=gray!45, dashed, line width=0.8pt,
      fill=gray!4,
      inner xsep=10pt, inner ysep=10pt,
      fit=(api)(pb)(llm)(ext)(val)(db)(audit)(result),
      label={[font=\footnotesize\itshape, text=gray!55, yshift=2pt]
             above:Local network perimeter}
    ] {};
  \end{scope}
\end{tikzpicture}}%
\caption{PharmaBatchDB AI local NLQ-to-SQL pipeline. The user submits a natural language
query; the FastAPI layer constructs a module-scoped prompt including the full schema and
routes it to the locally running Ollama inference engine. The model response is parsed
for SQL, validated by \texttt{sqlglot} AST inspection (SELECT-only, module-scoped table
check), and either executed against the read-only SQL Server instance or rejected.
Every query attempt---pass or fail---is written to the append-only audit log. All
inference, validation, and data access remain within the local network perimeter.}
\label{fig:architecture}
\end{figure}
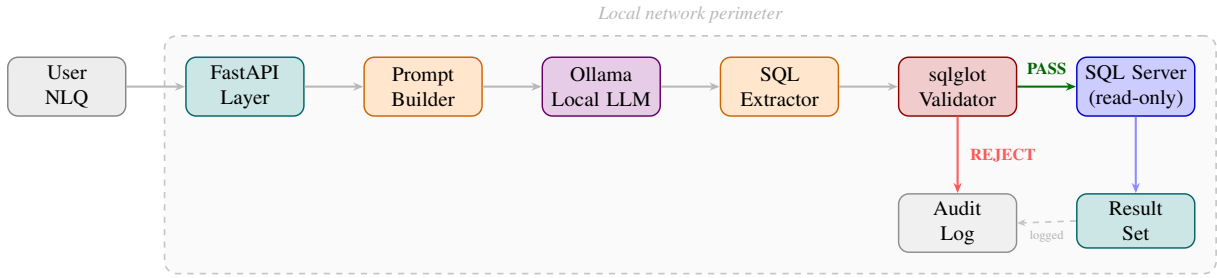

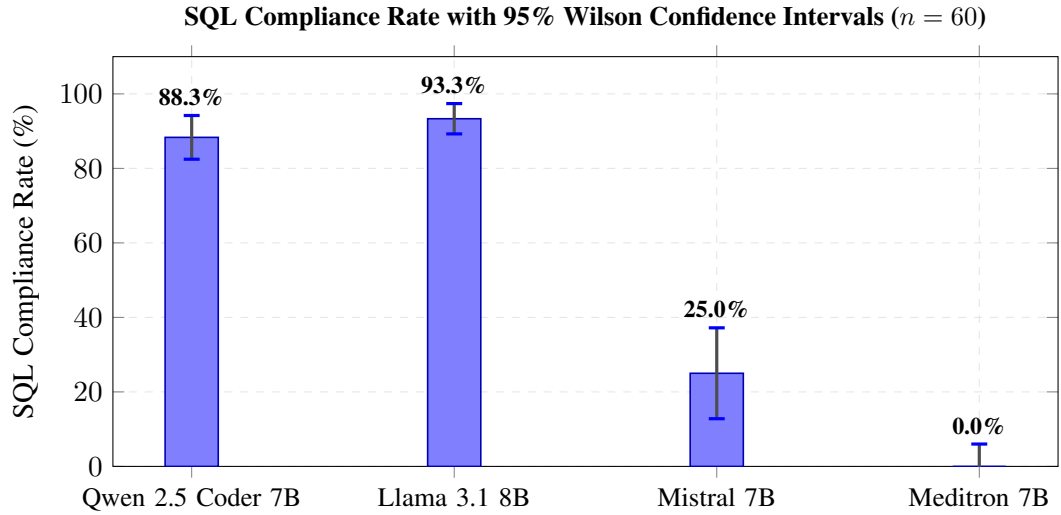
\begin{figure}[H]
\centering
\begin{tikzpicture}
\begin{axis}[
  ybar,
  bar width=20pt,
  width=0.88\textwidth,
  height=7cm,
  ylabel={SQL Compliance Rate (\%)},
  symbolic x coords={Qwen 2.5 Coder 7B, Llama 3.1 8B, Mistral 7B, Meditron 7B},
  xtick=data,
  xticklabel style={font=\small, text width=2.9cm, align=center},
  ymin=0, ymax=110,
  ytick={0,20,40,60,80,100},
  yticklabel style={/pgf/number format/fixed},
  error bars/y dir=both,
  error bars/y explicit,
  error bars/error bar style={line width=1.2pt, draw=black!70},
  error bars/error mark options={rotate=90, mark size=3pt, line width=1.2pt},
  grid=major,
  grid style={gray!20, dashed},
  title={SQL Compliance Rate with 95\% Wilson Confidence Intervals ($n=60$)},
  title style={font=\small\bfseries},
]
\addplot+[
  fill=blue!50!white,
  draw=blue!70!black,
  line width=0.6pt,
] coordinates {
  (Qwen 2.5 Coder 7B,  88.33) +- (10.53, 5.87)
  (Llama 3.1 8B,       93.33) +- (9.23,  4.07)
  (Mistral 7B,         25.00) +- (9.20, 12.20)
  (Meditron 7B,         0.00) +- (0.00,  6.00)
};
\node[font=\footnotesize\bfseries, anchor=south] at (axis cs:Qwen 2.5 Coder 7B,  94.2)  {88.3\%};
\node[font=\footnotesize\bfseries, anchor=south] at (axis cs:Llama 3.1 8B,       97.4)  {93.3\%};
\node[font=\footnotesize\bfseries, anchor=south] at (axis cs:Mistral 7B,         37.2)  {25.0\%};
\node[font=\footnotesize\bfseries, anchor=south] at (axis cs:Meditron 7B,         6.0)  {0.0\%};
\end{axis}
\end{tikzpicture}
\caption{SQL Compliance Rate (\%) for each model across all 60 benchmark questions.
Error bars show 95\% Wilson confidence intervals. Llama 3.1 8B and Qwen 2.5 Coder 7B
occupy the same performance tier; their confidence intervals overlap substantially
($p=0.529$, Fisher's exact). Mistral 7B shows a large, statistically significant drop
relative to both leading models ($p<0.001$). Meditron 7B produced 0\% compliance
across all questions in the standard-schema condition.}
\label{fig:compliance}
\end{figure}

\subsection{Performance by Manufacturing Module}
\label{sec:results-module}

Table~\ref{tab:bymodule} shows ROUGE-L, SQL Compliance, and Hallucination Rate
stratified by manufacturing module (20 questions each). Module complexity increases from
Batch (2 tables, simple foreign key joins) through MES (3 tables, equipment state
time-series) to CIP (2 tables, step-level process parameters with more complex join
conditions).

Mistral's compliance decreased sharply from Batch to the more complex modules: from
40.0\%\ (8/20) [21.9--61.3\%] on Batch to 15.0\%\ (3/20) [5.2--36.0\%] on MES and
20.0\%\ (4/20) [8.1--41.6\%] on CIP.
Even at per-module $n=20$, the direction of this decrease is practically meaningful. Llama achieved 95.0\%\ (19/20) [76.4--99.1\%] on Batch,
the highest of any model, and held at 95.0\%\ (19/20) [76.4--99.1\%] on MES and
90.0\%\ (18/20) [69.9--97.2\%] on CIP. Qwen's pattern matched Llama closely, with compliance of
90.0\%\ (18/20) [69.9--97.2\%] on both Batch and MES, dipping to 85.0\%\ (17/20) [64.0--94.8\%] on CIP.
At per-module $n=20$, these differences between Llama
and Qwen are within the confidence intervals and should not be over-interpreted.

\begin{table}[H]
\centering
\caption{Performance by manufacturing module (Batch/MES/CIP). SQL Compliance and
Hallucination Rate shown as \%\ (k/n) [95\% Wilson CI]. Per-module $n=20$.}
\label{tab:bymodule}
\resizebox{\textwidth}{!}{%
\begin{tabular}{llcccc}
\toprule
\textbf{Model} & \textbf{Module} & \textbf{ROUGE-L} & \textbf{SQL Compliance} & \textbf{Hallucination} \\
\midrule
\multirow{3}{*}{Qwen 2.5 Coder 7B}
  & Batch & 0.6705 & 90.0\%\ (18/20) [69.9--97.2\%]  & 10.0\%\ (2/20) [2.8--30.1\%] \\
  & MES   & 0.6166 & 90.0\%\ (18/20) [69.9--97.2\%]  & 10.0\%\ (2/20) [2.8--30.1\%] \\
  & CIP   & 0.6131 & 85.0\%\ (17/20) [64.0--94.8\%]  & 15.0\%\ (3/20) [5.2--36.0\%] \\
\midrule
\multirow{3}{*}{Llama 3.1 8B}
  & Batch & 0.5862 & \textbf{95.0\%\ (19/20) [76.4--99.1\%]} & \textbf{5.0\%\ (1/20) [0.9--23.6\%]} \\
  & MES   & 0.5401 & \textbf{95.0\%\ (19/20) [76.4--99.1\%]}  & \textbf{5.0\%\ (1/20) [0.9--23.6\%]} \\
  & CIP   & 0.5759 & 90.0\%\ (18/20) [69.9--97.2\%]  & 10.0\%\ (2/20) [2.8--30.1\%] \\
\midrule
\multirow{3}{*}{Mistral 7B}
  & Batch & 0.5517 & 40.0\%\ (8/20) [21.9--61.3\%]   & 60.0\%\ (12/20) [38.7--78.1\%] \\
  & MES   & 0.4612 & 15.0\%\ (3/20) [5.2--36.0\%]    & 85.0\%\ (17/20) [64.0--94.8\%] \\
  & CIP   & 0.4752 & 20.0\%\ (4/20) [8.1--41.6\%]    & 80.0\%\ (16/20) [58.4--91.9\%] \\
\midrule
\multirow{3}{*}{Meditron 7B}
  & Batch & 0.0000 & 0.0\%\ (0/20) [0.0--16.1\%] & 0.0\%$^{*}$ \\
  & MES   & 0.0143 & 0.0\%\ (0/20) [0.0--16.1\%] & 100.0\%\ (1/1)$^{*}$ \\
  & CIP   & 0.0000 & 0.0\%\ (0/20) [0.0--16.1\%] & 0.0\%$^{*}$ \\
\bottomrule
\end{tabular}
}
\footnotesize{$^{*}$Meditron 7B hallucination rate: 0.0\% reflects no SQL extracted
(model produced only narrative text); MES 100\%\ (1/1) reflects the single question
across all 60 where SQL was extractable but failed compliance validation. The HR
denominator for Meditron is the number of questions where SQL was actually extractable,
not the per-module total of 20. The SQL Compliance CI [0.0--16.1\%] at per-module
$n=20$ is the appropriate summary of Meditron's per-module output.}
\end{table}

\subsection{Performance by Question Difficulty}
\label{sec:results-difficulty}

Table~\ref{tab:bydifficulty} stratifies results by question difficulty (easy: $n=18$;
medium: $n=24$; hard: $n=18$), revealing distinct robustness patterns.

Qwen's most notable result was on medium-difficulty questions: 83.3\%\ (20/24)
[64.1--93.3\%] compliance, modestly lower than its easy performance (100.0\%,
18/18) [82.4--100.0\%]). Medium questions in this evaluation primarily involve
multi-table aggregations and conditional filtering; Qwen appears to have specific
failure modes in this region. Llama achieved a more consistent profile: 94.4\%\ (17/18)
[74.2--99.0\%] on both easy and hard, and 91.7\%\ (22/24) [74.2--97.7\%] on medium.
Qwen's hard-question compliance (83.3\%, 15/18) was lower than Llama's (94.4\%, 17/18),
a difference that warrants further investigation though it does not reach statistical
significance at these sample sizes. Mistral degraded monotonically from 44.4\% easy
to 5.6\% hard, with confidence intervals that leave no doubt about the directional trend.

\begin{table}[H]
\centering
\caption{Performance by question difficulty (easy/medium/hard). SQL Compliance and
Hallucination Rate shown as \%\ (k/n) [95\% Wilson CI]. Easy and Hard $n=18$; Medium $n=24$.}
\label{tab:bydifficulty}
\resizebox{\textwidth}{!}{%
\begin{tabular}{llcccc}
\toprule
\textbf{Model} & \textbf{Difficulty} & \textbf{ROUGE-L} & \textbf{SQL Compliance} & \textbf{Hallucination} \\
\midrule
\multirow{3}{*}{Qwen 2.5 Coder 7B}
  & Easy   & 0.7056 & \textbf{100.0\%\ (18/18) [82.4--100.0\%]}  & \textbf{0.0\%\ (0/18) [0.0--17.6\%]} \\
  & Medium & 0.6405 & 83.33\%\ (20/24) [64.1--93.3\%] & 16.67\%\ (4/24) [6.7--35.9\%] \\
  & Hard   & 0.5517 & 83.33\%\ (15/18) [60.8--94.2\%]  & 16.67\%\ (3/18) [5.8--39.2\%] \\
\midrule
\multirow{3}{*}{Llama 3.1 8B}
  & Easy   & 0.5700 & 94.44\%\ (17/18) [74.2--99.0\%]  & 5.56\%\ (1/18) [1.0--25.8\%] \\
  & Medium & 0.6048 & \textbf{91.67\%\ (22/24) [74.2--97.7\%]} & \textbf{8.33\%\ (2/24) [2.3--25.8\%]} \\
  & Hard   & 0.5150 & \textbf{94.44\%\ (17/18) [74.2--99.0\%]}  & \textbf{5.56\%\ (1/18) [1.0--25.8\%]} \\
\midrule
\multirow{3}{*}{Mistral 7B}
  & Easy   & 0.5320 & 44.44\%\ (8/18) [24.6--66.3\%]  & 55.56\%\ (10/18) [33.7--75.4\%] \\
  & Medium & 0.5051 & 25.00\%\ (6/24) [12.0--44.9\%]  & 75.00\%\ (18/24) [55.1--88.0\%] \\
  & Hard   & 0.4479 & 5.56\%\ (1/18) [1.0--25.8\%]    & 94.44\%\ (17/18) [74.2--99.0\%] \\
\midrule
\multirow{3}{*}{Meditron 7B}
  & Easy   & 0.0000 & 0.0\%\ (0/18) [0.0--17.6\%] & 0.0\% \\
  & Medium & 0.0119 & 0.0\%\ (0/24) [0.0--13.8\%] & 100.0\%\ (1/1)$^{\dagger}$ \\
  & Hard   & 0.0000 & 0.0\%\ (0/18) [0.0--17.6\%] & 0.0\% \\
\bottomrule
\end{tabular}
}
\footnotesize{$^{\dagger}$Meditron: 0.0\% hallucination cells indicate no SQL was extracted;
Medium 100\% reflects the single question across all 60 where SQL was extractable but invalid.}
\end{table}

\subsection{Statistical Analysis}
\label{sec:stats}

All proportional comparisons (SQL Compliance, Hallucination Rate) were tested using
Fisher's exact test, appropriate for binary outcomes at small $n$. ROUGE-L score
distributions were compared using the Wilcoxon signed-rank test (paired, two-tailed)
on per-question scores. Wilson 95\% confidence intervals are reported for all proportions
throughout this paper.

\textbf{SQL Compliance --- cross-model comparisons:}
The difference between Llama 3.1 8B (93.33\%, 56/60) and Qwen 2.5 Coder 7B
(88.33\%, 53/60) was not statistically significant (Fisher's exact $p=0.529$, OR=1.85).
This is a critical caveat for the interpretation of Table~\ref{tab:overall}: the
5.0 percentage point gap between the two leading models lies within the overlap
of their confidence intervals and cannot be treated as a meaningful performance
difference at $n=60$. Both models substantially outperformed Mistral 7B, and both
comparisons reached high significance (Llama vs Mistral: $p<0.001$, OR=42.00;
Qwen vs Mistral: $p<0.001$, OR=22.71). Meditron's 0\% compliance was clearly and substantially separated from all other models.

\textbf{ROUGE-L --- Wilcoxon signed-rank tests:}
All Wilcoxon signed-rank tests were computed using \texttt{scipy.stats.wilcoxon} \citep{virtanen2020scipy}
(\texttt{alternative="two-sided"}, \texttt{zero\_method="wilcox"}); questions were paired by question ID.
Qwen achieved significantly higher per-question ROUGE-L scores than Llama
(W=323.0, $p<0.001$) and Mistral (W=72.5, $p<0.001$); Llama also outperformed Mistral
(W=343.0, $p<0.001$). However, ROUGE-L reflects lexical similarity to single-author
reference queries as well as semantic correctness. The Qwen--Llama ROUGE-L difference
(0.6334 vs 0.5674) most plausibly reflects systematic differences in SQL stylistic
conventions---alias naming, ordering clauses, subquery structure---rather than a
meaningful difference in factual retrieval accuracy. Factual Consistency, which measures
result set overlap rather than query form, is therefore the primary quality metric in
this study (Qwen: 0.34; Llama: 0.30). FC differences between Qwen and Llama (0.34 vs 0.30) are reported descriptively.
The FC score distribution is right-skewed with many zero values (questions where
model-generated and reference result sets do not overlap), making parametric comparison
inappropriate. The point estimates should be treated as directional evidence only,
consistent with the non-significant compliance comparison. A zero-inflated model
would be required for valid inferential comparison of FC distributions, which is
beyond the scope of this proof-of-concept evaluation.

\textbf{Note on multiple comparisons:} All pairwise inferential tests reported above
should be interpreted as exploratory, consistent with the proof-of-concept sample size
of this study. No Bonferroni or Holm correction has been applied across pairwise
comparisons; the primary purpose of the tests is to characterise the magnitude and
direction of differences, not to provide confirmatory inference.

\section{Discussion}
\label{sec:discussion}

\subsection{The Meditron 7B Result and What It Suggests}
\label{sec:discussion-meditron}

The most significant result in this study is Meditron 7B's complete failure. Zero SQL
compliance across 60 questions, a ROUGE-L of 0.0048 that is essentially noise, and a
hallucination rate that, properly defined, applies to only 1 of the 60 questions
(the single case where SQL was actually extractable)---this pattern indicates a near-complete task failure rather than marginal underperformance. Given that Meditron was
specifically designed for biomedical tasks and the database schema is pharmaceutical in
nature, some degree of competitiveness with Mistral 7B, which shares a similar
architecture, might have been expected. The scale of the failure warrants careful
analysis rather than attribution to domain mismatch alone.

The most obvious explanation is the context window. Meditron 7B has a 2,048-token
context limit---a constraint inherited from its Llama 2 base and not expanded during
biomedical fine-tuning. The system prompt plus schema provided to all models consumes
approximately 1,800--2,000 tokens, which means that for Meditron, the schema description
was already pushing against or exceeding the available context before a single token of
the actual question was processed. Qwen and Mistral have 32,768-token windows; Llama 3.1
has 131,072. The module-by-module breakdown gives some indirect evidence: Meditron
produced no SQL at all for Batch and CIP, while for MES it extracted exactly one query
and that query was invalid. The MES prompts may have been structured in a way that
happened to leave just enough context for the model to generate something---but what it
generated was nonsense. This inconsistency suggests that context truncation alone does
not fully explain the observed failure pattern.

The second confounding factor is quantisation format. Meditron in the Ollama repository
uses Q4\_0, an older 4-bit scheme that applies uniform block quantisation without the
k-means grouping that Q4\_K\_M applies to improve precision on the most sensitive
weights. Q4\_K\_M is generally considered the better format for instruction following,
and the difference matters more for structured output tasks than for free text
generation. A Meditron Q4\_K\_M variant, if one were available, might perform somewhat
better---though the context window constraint would likely remain the dominant factor.

To disentangle these confounds, a follow-up experiment was run that truncated the schema
prompt to primary keys, foreign keys, and column names only (under 1,000 tokens) and
re-ran Meditron on the original 30-question set (B01--B10, M01--M10, C01--C10; 10 per module, stratified across all three difficulty levels) under this condition. The results are shown in the
``Meditron 7B (trunc.\ schema)'' row of Table~\ref{tab:overall} and are unambiguous.

SQL extraction improved dramatically: from 1/30 under the full schema to 28/30 under the
truncated prompt. Time to first token fell from 21,979~ms to 1,883~ms. Both changes are
strongly consistent with context-window saturation: the full schema prompt was
saturating the 2,048-token context window before the question was even processed,
leaving Meditron with no capacity to generate anything. Once the prompt was small enough
to fit, the model could at least produce SQL-like text.

What it could not do was produce \textit{valid} SQL. All 28 extracted queries failed
compliance validation, giving a 100\% hallucination rate among extractable outputs and
0\% SQL Compliance across all 30 questions---identical to the standard-schema run.
The SQL that Meditron generated was not simply incorrectly scoped; it referenced
non-existent tables, used MySQL syntax rather than T-SQL, and in several cases produced
fragments of SQL wrapped in clinical narrative text that the extractor pattern recognised
as SQL but the validator rejected immediately.

This result settles the question the standard-schema experiment left open. The context
window was the reason Meditron produced almost no output; it was not the reason the
output it produced was wrong. Even with room to generate, Meditron's SQL generation
capability on manufacturing schemas is absent. The most defensible interpretation is that
the Llama 2 base model, fine-tuned on biomedical text, had the SQL-generation skill
present in the base model diluted by a training distribution that contained very little
structured query generation at all---certainly nothing resembling T-SQL over manufacturing
schemas. Domain-specific biomedical pre-training, of the variety Meditron represents,
does not transfer to this task. These findings are consistent with \textit{skill dilution under domain-specific
fine-tuning}, sometimes described as catastrophic forgetting in the broader transfer
learning literature \citep{kirkpatrick2017catastrophic}, although the present experiment
cannot directly establish whether this reflects true forgetting of a previously acquired
capability or the absence of T-SQL generation capability in the Llama 2 base model's
training distribution. A plausible account is that fine-tuning on a corpus
dominated by biomedical narrative text (PubMed abstracts, clinical guidelines, case
reports) progressively attenuated the structured SQL generation behaviour present in the
Llama 2 base weights, as the model's parameters were redirected toward the new target
distribution. The result is consistent with a model that has absorbed biomedical knowledge at the cost
of SQL generation capability. Whether a Meditron variant built on Llama 3 (which has
a substantially stronger SQL baseline) would behave differently is a meaningful
follow-up question, but it is beyond the scope of this experiment.

\subsection{Llama and Qwen: A Difference Without a Statistically Significant Gap}

Llama 3.1 8B achieved the numerically highest SQL Compliance (93.33\%, 56/60) and
lowest Hallucination Rate (6.67\%, 4/60) across all 60 questions, and achieved the
highest compliance on both Batch and MES module queries (95.0\%, 19/20 each). These are desirable properties
for a pharmaceutical manufacturing NLQ system, where generating a non-compliant query is
not just a quality failure but a potential regulatory event if it causes incorrect data
to be displayed to someone making a manufacturing decision.

The statistical reality, however, is that Llama's compliance advantage over Qwen is not
significant even at $n=60$. The 5.0 percentage point difference---3 questions---is
within the overlap of their confidence intervals (Llama: 84.1--97.4\%; Qwen:
77.8--94.2\%), and Fisher's exact test confirms this ($p=0.529$). The more accurate framing is that both models are plausible candidates for deployment,
that Llama is numerically ahead on the compliance measure that matters most in regulated
contexts, and that a definitive separation between them would require a substantially
larger question set or a different evaluation design.

Qwen's higher ROUGE-L (0.6334 vs 0.5674, Wilcoxon $p<0.001$) and Factual Consistency
(0.34 vs 0.30) are similarly worth tempering. The ROUGE-L difference reflects, in part,
the reference author's SQL style---Qwen's output used alias conventions and subquery
structures more similar to the reference queries. A model can be valid, correctly scoped,
and return the right rows while producing SQL that looks quite different from a
single-author reference. The FC difference (0.34 vs 0.30) is the more operationally
meaningful comparison, but even there, the gap is modest and insufficient to support
a deployment decision on its own.

It is important to separate what the metrics indicate: SQL Compliance is a safety and
guardrail measure, indicating whether the model produced structurally valid, in-scope
queries; Factual Consistency is an answer-correctness measure, indicating whether those
queries returned the right data. The modest absolute FC values (0.34 for Qwen, 0.30 for
Llama) confirm that local 7B--8B models should not be deployed for unsupervised
operational decision-making, even at high compliance rates. Human review of both the
generated SQL and the result set remains necessary for any GxP-relevant query.

The Llama--Qwen comparison is statistically unresolved even at $n=60$; the Llama--Mistral
and Qwen--Mistral comparisons are not. On schemas with three or more tables and
non-trivial join conditions, both code-tuned models are substantially more reliable
than Mistral---and there the evidence is sufficiently large to inform practical model selection under the tested conditions.

The broader 2025--2026 model landscape provides additional context for this finding.
A pattern that appears consistently in benchmark comparisons is that generalist models
with large context windows and strong instruction-tuning can outperform domain-specific
models built on older base architectures when those specialist models lack the context
capacity to process complete domain schemas \citep{ascentcore2026benchmark}. The
Meditron result is consistent with this pattern in the present study: the biomedical
specialist could not fit the working schema in context, while the general-purpose Llama
model processed it without difficulty. This observation should not be over-generalised
beyond the evidence available here, but it does suggest that practitioners evaluating
model selection for regulated environments should weigh context window capacity alongside
domain pre-training when assessing candidate models for structured schema tasks.

\subsection{Qwen 2.5 Coder 7B: The Curious Medium-Difficulty Problem}

Qwen's profile deserves a second look beyond the aggregate figures. The highest ROUGE-L
and FC, combined with 100\% compliance on easy questions, suggest a model that
is genuinely well-calibrated for SQL generation at the simpler end of the difficulty
spectrum. At $n=60$, however, a more nuanced picture emerges: Qwen achieved 83.3\%
compliance on both medium (20/24, CI 64.1--93.3\%) and hard (15/18, CI 60.8--94.2\%)
questions, while Llama held at 94.4\% on hard (17/18, CI 74.2--99.0\%). The medium
questions in this evaluation tend to involve multi-table aggregations with conditional
filtering---a category that may expose specific failure modes in Qwen's training
distribution that do not appear under simpler query types.

This anomaly remains difficult to interpret and should be treated cautiously. The most
defensible recommendation for practitioners is to treat medium-complexity multi-table
aggregations as the highest-risk query category for Qwen and implement downstream
validation accordingly. The SQL compliance layer in PharmaBatchDB AI is designed
precisely for this purpose, and the difficulty-stratified results here provide an
empirical basis for configuring where additional scrutiny is warranted.

\subsection{Mistral 7B and Schema Complexity Sensitivity}

Mistral 7B's trajectory is the clearest demonstration of schema complexity sensitivity
in these results. From 40.0\% compliance (8/20) on two-table Batch queries to 15.0\%
(3/20) on MES and 20.0\% (4/20) on CIP---at $n=20$ per module, the confidence intervals
remain wide, but the direction is unambiguous. Whether this reflects a fundamental limitation of Mistral's SQL training
or a sensitivity to the specific join vocabulary of the PharmaBatchDB schema is
something this data cannot determine. The distinction matters because the remediation
differs: a fundamental SQL training limitation would persist on any complex schema,
while a sensitivity specific to PharmaBatchDB's join vocabulary---its use of implicit
foreign key conventions and manufacturing-specific column naming---might be addressable
through schema annotation or a richer system prompt. A replication against a second
schema with different join depth and vocabulary would separate these cases; the
SAP PM module data structure would be a natural candidate given its prevalence in
mid-to-large pharmaceutical sites.

\subsection{Implications for GxP-Aligned Local Deployment}

This study provides some evidence that a fully local, GxP-aligned NLQ pipeline is
feasible on hardware that is not prohibitively expensive. Inference rates of 3.2 to 4.4
tokens per second under CPU-primary conditions yield practical time-to-first-token values
of roughly 11 to 50 seconds---acceptable for analytical query workflows involving batch
record review or equipment monitoring, though clearly not suitable for interactive
dashboards expecting sub-second responses.

Rather than asserting alignment with regulatory guidance, Table~\ref{tab:gxp} maps the
specific requirements from the January 2026 EMA/FDA joint guidance
\citep{fdaema2026aiguidance} and FDA 21 CFR Part 11 to the system features of
PharmaBatchDB AI, with the implementation evidence for each. Readers can assess the
adequacy of each mapping themselves.

\begin{table}[H]
\centering
\caption{Mapping of GxP-Relevant Design Expectations to System Features and
Implementation Evidence in PharmaBatchDB AI.}
\label{tab:gxp}
\begin{tabularx}{\textwidth}{>{\raggedright\arraybackslash}p{4.2cm}
                              >{\raggedright\arraybackslash}p{3.8cm}
                              >{\raggedright\arraybackslash}X}
\toprule
\textbf{Regulatory Requirement} & \textbf{System Feature} & \textbf{Implementation Evidence} \\
\midrule
FDA 21 CFR Part 11 --- Electronic audit trails \citep{fda21cfr11}
  & Append-only query log
  & \texttt{audit\_log.py} records every query with timestamp, authenticated user,
    generated SQL, execution time, model, TPS, and TTFT; log is append-only and
    not modifiable by the LLM or API layer \\[6pt]
EU AI Act Art.\ 13 (Transparency to deployers) \citep{eu2024aiact}
  & Audit log with AI attribution; generated SQL displayed before execution
  & Every query is logged with model identity, TPS, TTFT, and generated SQL;
    the analyst sees the AI-generated SQL before it is executed, ensuring
    no AI output reaches operational use without human interpretation \\[6pt]
EMA/FDA 2026 Guidance --- Human oversight \citep{fdaema2026aiguidance}
  & No direct DB access; results reviewed before operational use
  & Read-only SQL user (\texttt{ai\_readonly}); generated SQL is displayed to
    the human analyst before execution; no autonomous action pathway exists \\[6pt]
GxP Annex 11 --- Data integrity \citep{eu_annex11}
  & SELECT-only AST validation; DML/DDL blocked
  & All INSERT, UPDATE, DELETE, and DDL statements are blocked at the
    \texttt{sqlglot} T-SQL AST validation layer before reaching the database engine;
    only \texttt{SELECT} statements are permitted to execute \\[6pt]
PIC/S GMP --- Audit trail review \citep{pics_gmp}
  & Per-query logging of model identity, TPS, and TTFT
  & Supports post-hoc investigation of AI-generated queries; logged fields allow
    reconstruction of which model produced which SQL under which conditions \\
\bottomrule
\end{tabularx}
\end{table}

A practical consideration that the latency figures in this study surface is what may be
termed the \textit{performance tax} of on-premises deployment. At 3.2 to 4.4 tokens per
second on consumer-grade CPU hardware, analysts accustomed to cloud-based AI interfaces
will notice the difference; GPU-accelerated deployments report substantially higher
throughput for comparable model sizes \citep{ascentcore2026benchmark}. The risk this
creates in regulated environments deserves explicit acknowledgement: if the local system
is noticeably slower than freely available cloud alternatives, some users may route
sensitive manufacturing queries through non-compliant channels. This \textit{shadow AI}
problem---the informal use of consumer AI tools to process data that should remain within
the organisational perimeter---is a well-documented adoption failure mode in regulated
industries \citep{puthal2025shadowai}. Hardware investment to reduce this performance gap
should therefore be framed not as a performance optimisation but as a
compliance-enabling expenditure: it removes a key motivation for shadow AI use.

A second, less obvious implication of the synthetic data methodology concerns its utility
as a \textit{pre-certification pathway} for regulated deployments. Validation of AI
systems under GxP frameworks---particularly Annex~11 and FDA 21~CFR Part~11---requires
documented testing before the system processes real regulated data. A synthetic database
that faithfully replicates the schema, cardinality, and query complexity of the production
system allows the full benchmark and validation cycle to be completed independently of the
regulated data environment. In practice, a substantial portion of the technical validation evidence required for a
change-control submission can be assembled before the system touches production batch
records. The benchmark question set, reference queries, and evaluation
harness developed for this study constitute precisely the artefacts that would form the
technical component of such a submission package, and this reuse pathway may lower the
adoption barrier for sites that are hesitant to engage with AI tools before they are
fully validated.

\subsection{Limitations}
\label{sec:limitations}

The sample size of 60 questions per model is the most consequential methodological
constraint in this study, and it has been discussed at length in Section~\ref{sec:stats}.
To restate the key point: this evaluation can distinguish large performance gaps with
reasonable confidence, but the fine-grained comparison between Llama and Qwen is not
statistically resolved at $n=60$.

The Q4\_K\_M quantisation used for three of the four models represents a real quality
trade-off relative to Q8\_0 inference, and the CPU-primary execution environment is
not comparable to GPU-accelerated deployments. Both constraints reflect real-world
conditions for many potential adopters, which partly motivates their choice, but they
limit direct comparability with published benchmarks that used GPU hardware or
higher-precision quantisation.

The synthetic nature of PharmaBatchDB is perhaps the most significant practical
limitation. Synthetic data tends to be more internally consistent and more closely
aligned with the schema definition than real operational databases, where column names
carry historical accidents, data quality varies, and implicit domain knowledge is
required to formulate semantically correct queries. How much this matters in practice cannot be assessed without access to a real
pharmaceutical manufacturing database under appropriate data governance agreements.
Performance on real production schemas is likely lower than on synthetic data, where
naming conventions are consistent and implicit domain knowledge is not required. The
compliance rates reported for Llama (93.33\%) and Qwen (88.33\%) should therefore be
treated as plausible upper bounds rather than direct estimates of operational performance,
and any deployment against a production schema should be preceded by schema-specific
evaluation using real or representative queries.

The reference queries were authored by the same investigator who designed the schema.
This introduces bias in ROUGE-L scores---the models that write SQL in a style similar to
the investigator's will appear to perform better than models that write semantically
equivalent but stylistically different queries. This bias partly motivates the adoption
of Factual Consistency as the primary metric, but FC is not immune to authorship effects
either: if the reference query is subtly miscalibrated, FC will measure conformity to
that miscalibration. A follow-up study using a panel of reference SQL authors, including
QA specialists, would substantially reduce this risk.

This study did not include cloud-hosted frontier models (GPT-4o, Claude, Gemini) as
comparison baselines. This exclusion was deliberate---the study evaluates the feasibility
of the local-only deployment scenario---but it limits interpretability for practitioners
who need to quantify the performance cost of data-sovereignty constraints relative to
cloud alternatives.

Factual Consistency was assessed using automated Jaccard similarity on result row sets.
No human expert rating of semantic correctness was performed. A panel of QA specialists
or pharmacovigilance database experts independently evaluating the correctness of
model-generated queries would provide a stronger validity criterion than the automated
FC metric alone.

This evaluation used a single synthetic schema (PharmaBatchDB). The module-level and
difficulty-stratified results cannot be assumed to generalise to other pharmaceutical
schemas with different join depths, column vocabularies, or domain naming conventions.
The model ranking observed here---particularly the fine-grained comparison between Llama
and Qwen---may differ on alternative schemas.

Each model was run once per question at temperature 0.1. Near-deterministic output was
the goal, but results are not guaranteed to be perfectly reproducible across separate
Ollama runs due to floating-point non-determinism at the hardware level. Repeated
sampling trials would provide confidence intervals on per-question model behaviour.

Question difficulty labels (easy/medium/hard) were assigned by a single investigator
prior to any model evaluation, based on the number of tables joined, the presence of
aggregation or window functions, and the specificity of filter conditions. No inter-rater
reliability measure (e.g., Cohen's kappa) was computed. The difficulty-stratified results
in Table~\ref{tab:bydifficulty} should therefore be treated as exploratory: the classification
is operationally defined and internally consistent, but its reproducibility across
independent raters has not been established.

The zero-shot prompting strategy used here evaluates baseline schema comprehension
without prompt engineering. Performance may change substantially with few-shot examples,
schema annotations, or retrieval-augmented schema presentation. The reported compliance
rates should not be assumed to represent the ceiling of achievable performance.

\section{Conclusion}
\label{sec:conclusion}

This paper set out to evaluate whether locally deployed 7B--8B parameter LLMs can
provide reliable NLQ-to-T-SQL generation over pharmaceutical manufacturing data on
consumer-grade hardware. The findings demonstrate that performance varies substantially
across models: the data reliably distinguish broad performance tiers, while fine-grained
within-tier rankings remain statistically unresolved at the current sample size.

Two models produced genuinely useful results. Llama 3.1 8B showed numerically higher
compliance (93.33\%, 56/60) than Qwen 2.5 Coder 7B (88.33\%, 53/60), though this
difference did not reach statistical significance at $n=60$ (Fisher's exact $p=0.529$),
and a stronger recommendation between the two awaits a larger evaluation. Both models
substantially outperformed Mistral 7B on complex multi-table schemas---that comparison
is statistically secure and practically meaningful. For organisations prioritising
reliability on structured output pipelines in regulated contexts, either Llama or Qwen
is a reasonable starting point; Llama's numerically better compliance profile and its
100\% Batch module performance suggest it is the lower-risk default when a validation
layer is not present to catch failures, while Qwen's higher Factual Consistency may
appeal where query quality on well-constrained questions is the primary concern.

Meditron 7B's complete failure is the finding most likely to generate discussion among
practitioners who assume that biomedical pre-training should confer some advantage in
pharmaceutical contexts. The truncated-schema follow-up reported here resolves the central
confound: the 2,048-token context window was suppressing SQL extraction in the
standard-schema condition, but once the prompt was small enough to fit, Meditron still
produced no valid SQL (0/60 compliance, 100\% hallucination rate among extractable
outputs). The failure is not purely architectural---context overflow was masking an
underlying generation capability deficit. The results are consistent with skill dilution
under domain-specific fine-tuning: biomedical pre-training on the Llama 2 base does not
appear to transfer to T-SQL generation over manufacturing schemas, a finding that should
inform model selection for NLQ systems in regulated environments.
Mistral 7B's compliance decrease on multi-table schemas suggests it is unsuitable under
zero-shot conditions and the schema complexity tested here, despite performing adequately
on simple two-table queries.

For practitioners evaluating whether local LLM deployment is worth pursuing, the
feasibility evidence here is reasonably encouraging: a fully local, auditable NLQ
pipeline with GxP-aligned design features---append-only logging, read-only database
access, AST-level SQL validation---is implementable on consumer hardware without
specialist infrastructure. The modest absolute FC values confirm, however, that human
review remains necessary and that current 7B--8B models are best understood as
query-drafting tools rather than autonomous query generators.

Future directions that would most productively extend this work follow directly from
the findings and limitations above, rather than from generic NLP research trends:

\begin{itemize}[noitemsep]
  \item Validation against real pharmaceutical ERP and LIMS export data, using
  production SQL query logs as reference gold standard, to assess whether the
  performance profiles observed on synthetic data transfer to operational schemas
  with real data-quality variability.

  \item Evaluation of Meditron variants with extended context windows---specifically
  Meditron variants fine-tuned from Llama 3 rather than Llama 2---to disentangle
  the contribution of domain knowledge from the context window architectural constraint
  identified in this study.

  \item Multi-author reference SQL construction involving QA specialists who are
  blind to the schema design, to reduce the authorship bias that affects both ROUGE-L
  interpretation and FC calibration in single-author evaluations.

  \item Integration with LIMS and ERP schemas beyond the seven-table PharmaBatchDB
  structure---SAP PM modules and Veeva Vault quality schemas are natural next targets,
  given their prevalence in mid-to-large biopharmaceutical sites.

  \item Evaluation of few-shot prompting strategies using regulatory query examples
  authored by qualified QA specialists, to assess how much of the zero-shot compliance
  gap between models can be closed through prompt engineering alone.

  \item A practitioner survey of biopharmaceutical professionals to establish the
  adoption envelope that complements this technical benchmark---specifically, whether
  practitioners find the hallucination rates observed here acceptable for GxP-supervised
  use, and what barriers (data sovereignty, IT infrastructure, validation overhead)
  most strongly predict local versus cloud AI preference in manufacturing settings.
\end{itemize}

\section*{Data Availability Statement}

The evaluation harness, benchmark question set (\texttt{benchmark\_questions.json}),
hardware profile (\texttt{hardware\_profile.md}), and anonymised benchmark results
(\texttt{benchmark\_results.json}) are available in the project repository at
\url{https://github.com/sbhetwal11/Benchmarking-Local-Large-Language-Models-for-Natural-Language-to-SQL}.
The synthetic database schema (\texttt{schema.md}) and data generation scripts are
included.

\section*{Authors' Contributions}

\textbf{SB} (Sagar Bhetwal): Conceptualisation, Methodology, Software, Data Curation, Formal
Analysis, Writing --- Original Draft, Writing --- Review and Editing.
\textbf{RB} (Rajan Bastakoti): Conceptualisation, Methodology, Writing --- Review and Editing,
Supervision.
\textbf{NA} (Nirajan Acharya): Writing --- Review and Editing, Validation, Investigation.
\textbf{GG} (Gaurav Gupta): Writing --- Review and Editing, Validation, Investigation.

\section*{Conflicts of Interest}

The authors declare no conflicts of interest.

\section*{Acknowledgements}

None declared.

\bibliographystyle{abbrvnat}
\bibliography{references}

\end{document}